%% file: main.tex
\documentclass[10pt,twocolumn,letterpaper]{article}

\usepackage[pagenumbers]{cvpr} 

\input{preamble}

\definecolor{cvprblue}{rgb}{0.21,0.49,0.74}
\usepackage[pagebackref,breaklinks,colorlinks,citecolor=cvprblue]{hyperref}

\def\paperID{***}
\def\confName{3DV\xspace}
\def\confYear{2026\xspace}

\title{KaoLRM: Repurposing Pre-trained Large Reconstruction Models \\ for Parametric 3D Face Reconstruction}

\author{Qingtian Zhu\textsuperscript{1}\footnotemark[1] ,
Xu Cao\textsuperscript{2}, 
Zhixiang Wang\textsuperscript{2},
Yinqiang Zheng\textsuperscript{1},
Takafumi Taketomi\textsuperscript{2}\\
\textsuperscript{1} The University of Tokyo\quad \textsuperscript{2} CyberAgent \\
}

\begin{document}

\input{sec/teaser}
\footnotetext[1]{Work done during Q. Zhu’s internship at CyberAgent AI Lab.}
\input{sec/0_abstract}    
\input{sec/1_introduction}
\input{sec/2_related}

\input{sec/3_method}
\input{sec/4_experiment}
\input{sec/5_discussion}
\input{sec/6_conclusion}
\clearpage

{
    \small
    \bibliographystyle{ieeenat_fullname}
    \bibliography{main}
}

\input{sec/X_suppl}

\end{document}

%% file: preamble.tex
\usepackage[dvipsnames]{xcolor}

\usepackage{minted}

\usepackage{booktabs}
\usepackage{multirow}
\usepackage{colortbl}

\newcommand{\textbfit}[1]{\textbf{\textit{#1}}}
\newcommand{\ourmethod}{KaoLRM\xspace}
\newcommand{\facemask}{\mathbf{m}}

\newcolumntype{g}{>{\columncolor{gray!20}}c}

\newcommand{\lrm}{LRM~\cite{hong2024lrm}\xspace}

%% file: sec/teaser.tex
\twocolumn[{
\renewcommand\twocolumn[1][]{#1}
\maketitle
\centering
\thispagestyle{empty}
\vspace{-10pt}
\includegraphics[width=\linewidth]{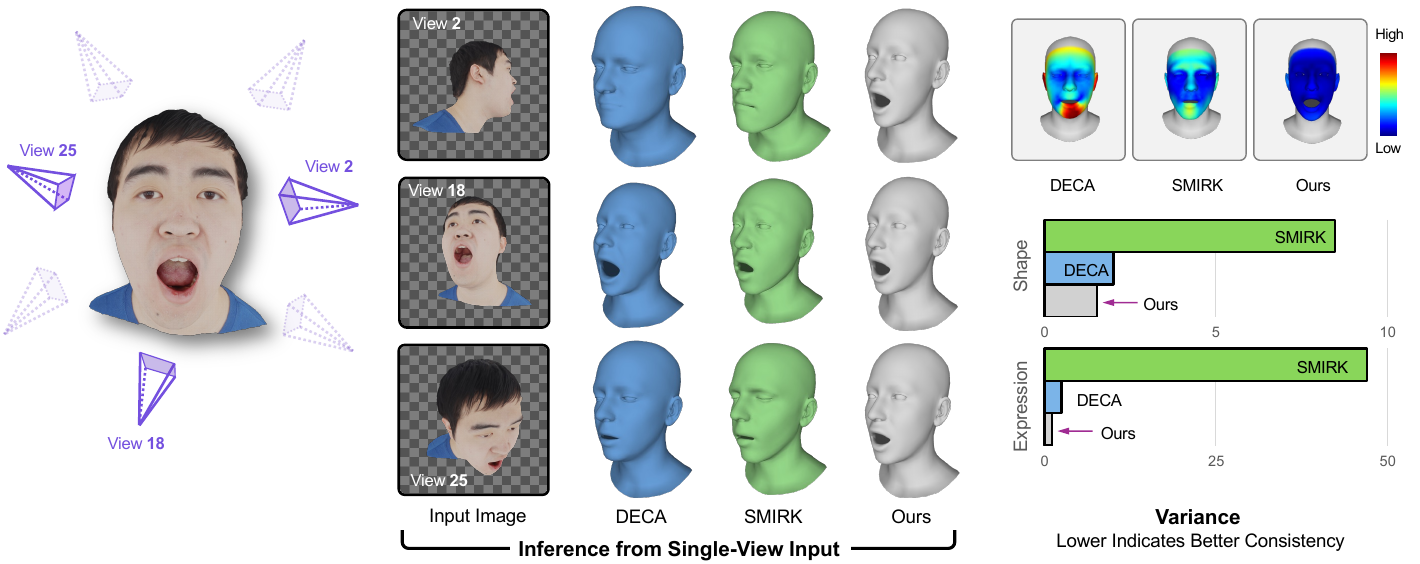}
\vspace{-2em}
\captionof{figure}{\ourmethod predicts FLAME parameters from a \emph{single-view} facial image and yields more faithful and \emph{consistent} predictions for the same subject captured from different viewpoints.
We validate the consistency qualitatively through visual inspection of the reconstructed FLAME meshes and quantitatively by measuring the variance of the FLAME parameters across the views. The methods for reference are DECA~\cite{feng2021learning} and SMIRK~\cite{retsinas20243d}.
}
\label{fig:teaser}
\vspace{3em}
}]

%% file: sec/0_abstract.tex
\begin{abstract}

We propose KaoLRM to re-target the learned prior of the Large Reconstruction Model (LRM) for parametric 3D face reconstruction from single-view images.
Parametric 3D Morphable Models (3DMMs) have been widely used for facial reconstruction due to their compact and interpretable parameterization, yet existing 3DMM regressors often exhibit poor consistency across varying viewpoints.
To address this, we harness the pre-trained 3D prior of LRM and incorporate FLAME-based 2D Gaussian Splatting into LRM's rendering pipeline.
Specifically, KaoLRM projects LRM’s pre-trained triplane features into the FLAME parameter space to recover geometry, and models appearance via 2D Gaussian primitives that are tightly coupled to the FLAME mesh.
The rich prior enables the FLAME regressor to be aware of the 3D structure, leading to accurate and robust reconstructions under self-occlusions and diverse viewpoints.
Experiments on both controlled and in-the-wild benchmarks demonstrate that KaoLRM achieves superior reconstruction accuracy and cross-view consistency, while existing methods remain sensitive to viewpoint variations.
The code is released at \url{https://github.com/CyberAgentAILab/KaoLRM}.

\end{abstract}
\vspace{-10pt}

%% file: sec/1_introduction.tex
\section{Introduction}
Inferring parametric 3D Morphable Models (3DMMs) from single-view facial images is a long-standing challenge in computer vision and graphics.
Due to the compact and interpretable parametric representation, 3DMMs provide intuitive control over identity, expression, and pose, making them crucial for applications such as animation, virtual reality, and telepresence~\cite{thies2018facevr}.

Training single-view 3DMM regressors with annotated parameters or explicit 3D supervision requires paired data, which are often expensive to obtain.
To alleviate the reliance on such annotations, many existing methods~\cite{feng2021learning,danvevcek2022emoca, retsinas20243d} adopt self-supervised, analysis-by-synthesis training paradigms, where the reconstructed 3DMM is re-rendered from the input viewpoint and compared with the input images using a photometric loss.
However, this paradigm lacks cross-view supervision, making the prediction sensitive to viewpoint changes at test time.
Without cross-view constraints, the model relies primarily on 2D visual features to regress 3DMM parameters, rather than inferring the 3D structure in the image.
Consequently, when viewpoint changes and key facial components become self-occluded, the reconstruction deteriorates rapidly, as shown in \cref{fig:teaser}.

Recent advances in 3D foundation models~\cite{hong2024lrm,wang2024dust3r,wang2025vggt} have marked a paradigm shift in 3D vision.
Rather than relying heavily on inductive biases and task-specific architectures, these models exploit massive amounts of diverse 3D data and high-capacity neural networks to learn general-purpose priors for 3D reconstruction.
For example, LRMs~\cite{hong2024lrm} reconstruct implicit 3D representations (\eg, triplane-based radiance fields~\cite{mildenhall2020nerf,chan2022efficient}) with cross-view photometric supervisions that are robust to viewpoint changes and missing features. 
However, these implicit representations are not directly suitable for downstream applications that require structured, parameterized control, leaving a gap between general-purpose reconstruction models and task-specific parametric modeling.

\begin{figure}
    \centering
    \includegraphics[width=\linewidth]{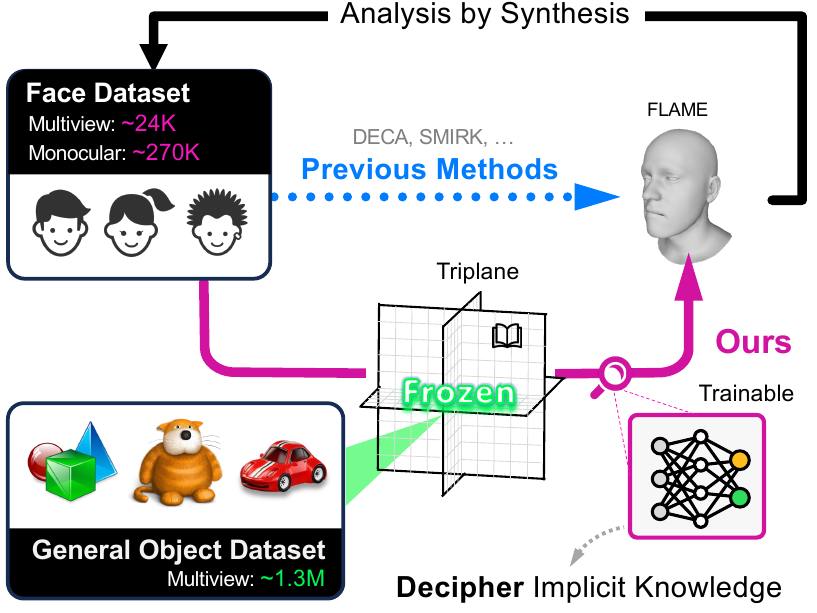}
    \caption{Motivation of \ourmethod. Instead of learning 3DMM regressors from scratch using multi-view face datasets that are expensive to scale,
    we leverage the learned 3D prior of LRM~\cite{hong2024lrm} (\eg, tri-plane features), allowing \ourmethod to be trained with a moderate amount of data.}
    \label{fig:main_idea}
\end{figure}

To address this discrepancy, we propose \textbf{\ourmethod}\footnote{``Kao" is derived from the Japanese word for ``face".} to adapt the pre-trained prior of LRM for parametric face reconstruction. 
\Cref{fig:main_idea} illustrates our key motivation.
Leveraging LRM's priors provides two main benefits.
First, the rich 3D prior encoded in LRM captures 3D structure in images more effectively, improving the robustness to invisible facial components under viewpoint changes.
Second, pre-training on a large-scale multi-view general object dataset reduces the reliance on large-scale multi-view facial datasets.

Specifically, we incorporate a FLAME decoder into LRM to project the pre-trained triplane features into the FLAME parameter space, enabling parametric facial geometry prediction.
To support cross-view photometric supervision, we model appearance using surface-bounded 2D Gaussian primitives derived from the FLAME mesh, and introduce a binding loss to enforce tight coupling between geometry and appearance.
This coupling prevents the model from overfitting to appearance while ignoring geometric accuracy.
Experimental results demonstrate that KaoLRM achieves superior reconstruction accuracy and improved cross-view consistency compared to prior methods, validating the effectiveness of transferring LRM’s pre-trained prior from general objects to parametric face models.

In summary, our contributions are threefold:
\begin{itemize}
    \item We introduce \ourmethod, a framework that re-targets the pre-trained knowledge of LRM for parametric face reconstruction.
    \item We bridge the gap between LRM's implicit 3D representations and FLAME parameters by integrating a parametric analysis-by-synthesis pathway into LRM’s rendering framework.
    \item \ourmethod outperforms previous methods regarding reconstruction quality and view robustness while maintaining competitive performance in in-the-wild scenarios. 
\end{itemize}

%% file: sec/2_related.tex
\section{Related Work}

\subsection{Large Reconstruction Model}

As the prevailing trend in both neural network architectures and dataset sizes continues to move towards larger scales, one notable outcome in the specific domain of 3D reconstruction is the emergence of the Large Reconstruction Model (LRM)~\cite{hong2024lrm}.
The underlying philosophy of LRM is to incorporate as little inductive bias as possible, \eg, the camera geometry usually adopted for unprojection, so that the model can learn representations and mappings directly from data, potentially surpassing the constraints of traditional, hand-crafted geometric assumptions.
As a result, LRM can regress the triplane-based radiance field directly from a single image in a completely feed-forward manner.

Following LRM, a number of methods have emerged that further investigate and expand upon different facets of this paradigm. 
LGM~\cite{tang2024lgm} and GS-LRM~\cite{zhang2024gs} shift the target of regression from radiance fields to pixel-aligned 3D Gaussians;
PF-LRM~\cite{wang2024pflrm} studies the power of LRM in joint pose and shape prediction; 
MeshLRM~\cite{wei2024meshlrm} re-targets a NeRF LRM to directly generate meshes with higher quality;
LRM-Zero~\cite{xie2024lrm} relieves the data reliance of LRM by employing synthetic data for training;
LVSM~\cite{jin2025lvsm} further extends the paradigm to novel view synthesis.

Specifically regarding LRM approaches to digital human avatars, we observe the emergence of methods such as Human-LRM~\cite{weng2024template} and FaceLift~\cite{lyu2024facelift}, both of which rely on multi-view diffusion models to generate pseudo multi-view inputs and employ LRM-based models to perform reconstruction.
However, we observe that these methods produce generic 3D geometric representations, such as point clouds or meshes, which are not readily applicable to downstream tasks like animation, where parametric models such as SMPL~\cite{loper2015smpl} or FLAME~\cite{li2017learning} are typically preferred.
In this paper, we opt to explore the possibility of re-targeting LRM to a compact 3D representation, namely a parametrized 3DMM, to achieve a more reasonable balance between generalizability and practicality.

\subsection{3DMM Regression}

3DMM~\cite{egger20203d} refers to parametric and statistical models of 3D face shape (and optionally appearance) built from real 3D scans.
The original 3DMM~\cite{BlanzVetter1999} is a typical example where Principal Component Analysis (PCA) is applied to a dataset of registered 3D faces to model variations in both shape and texture.
Since then, many specialized 3DMMs have been developed, \eg, BFM~\cite{paysan2009face}, LSFM~\cite{booth20163d}, and FLAME~\cite{li2017learning}.
It is also worth noting that in addition to conventional 3DMMs, some works~\cite{giebenhain2023learning,martinez2024codec} use a learnable neural network to achieve decoding from latent to mesh, thereby enabling parametric modeling with neural networks.
In this paper, we exclusively employ FLAME as the target parametric model.

Though there are attempts with direct 3D supervision~\cite{zielonka2022towards,zhang2023accurate}, considering the lack of high-quality 3D ground-truth annotations, most of the methods are trained in a self-supervised manner, by enforcing various types of consistency loss.
MoFA~\cite{tewari2017mofa,tewari2018high} introduces a model-based deep autoencoder that enables unsupervised learning via differentiable rendering. 
This analysis-by-synthesis paradigm was subsequently advanced to achieve fast reconstruction~\cite{tewari2018self} and to learn the generative face model itself from videos~\cite{tewari2019fml}.
Most of the subsequent 3DMM regressors focus on the design of the synthesis procedure including the choice of appearance modeling methods and the formulation of loss functions—particularly, the type of consistency enforced between the synthesized and observed images, such as photometric, perceptual, or semantic consistency.
Following this line of research, RingNet~\cite{sanyal2019learning} relies on the alignment of facial landmarks and the shape consistency between images of the same identity.
DECA~\cite{feng2021learning} further solidifies the analysis-by-synthesis approach for FLAME by enforcing robust photometric consistency.
Meanwhile, for real-time applications, 3DDFA-V2~\cite{guo2020towards} utilizes a lightweight backbone to regress 3DMM parameters and introduces a meta-joint optimization strategy and a 3D-aided short-video synthesis method.
Recent works have integrated perceptual constraints to enhance the realism of facial dynamics in 3D reconstruction. 
For instance, EMOCA~\cite{danvevcek2022emoca} exploits emotion-consistency during training to improve the expressiveness and robustness of the reconstructed avatars under varying emotions and facial expressions.
In addition, SPECTRE~\cite{filntisis2023spectre} leverages a lipread loss to improve the reconstruction of mouth movements.
To mitigate the impact of domain gap on optimization, SMIRK~\cite{retsinas20243d} adopts an analysis-by-neural-synthesis approach that employs a generative neural rendering module to replace the traditional differentiable rendering.
SHeaP~\cite{schoneveld2025sheap} learns FLAME mesh reconstruction and tracking via rigged head motions and use the combination of Gaussian prototypes based on face-aligned features to model the appearance.

We position our method \ourmethod within the analysis-by-synthesis paradigm, similar to DECA~\cite{feng2021learning}, EMOCA~\cite{danvevcek2022emoca}, and SMIRK~\cite{retsinas20243d}. 
Distinctively, \ourmethod leverages the pre-trained LRM prior~\cite{hong2024lrm} to obtain 3D-aware features for FLAME parameter estimation, and models appearance by converting LRM-generated triplane features into a Gaussian-based representation.

%% file: sec/3_method.tex
\begin{figure*}[t]
\centering
\includegraphics[width=\linewidth]{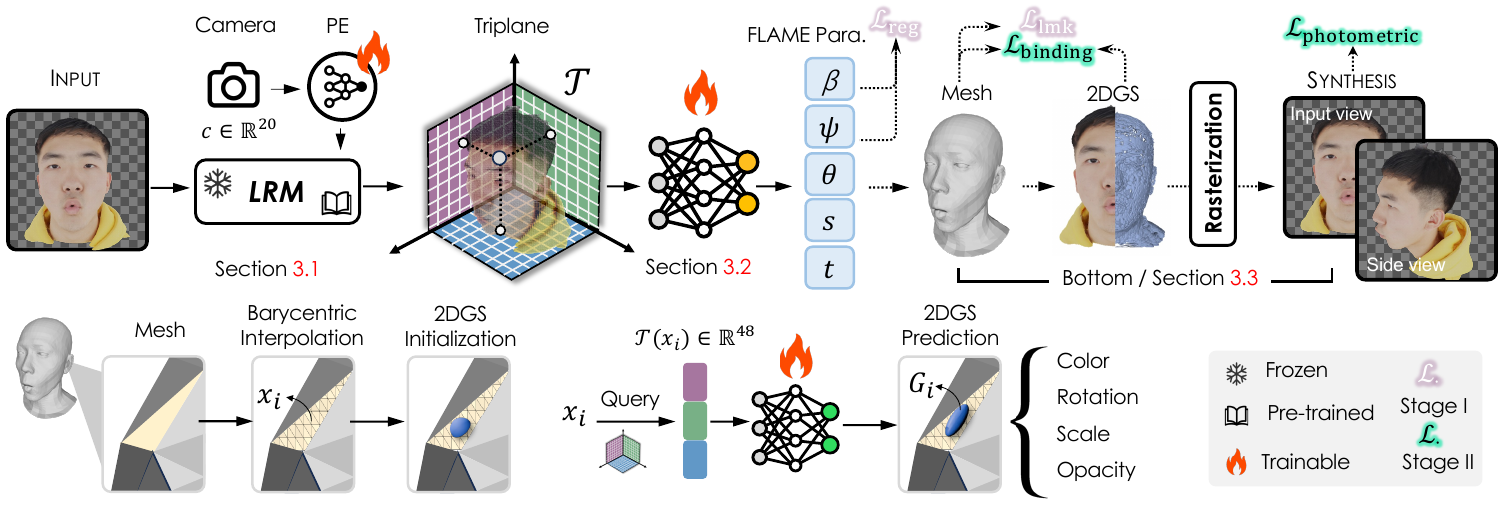}
\caption{Overall architecture of \ourmethod. 
\textbf{Top}: \ourmethod leverages the pre-trained image-to-triplane transformer of \lrm, and trains an additional network to predict FLAME parameters from the triplane features. The resulting FLAME mesh is then converted to 2D Gaussian primitives for synthesizing input-view or novel-view images. 
\textbf{Bottom}: After obtaining the FLAME mesh, we sample points via barycentric interpolation to initialize the centers of 2D Gaussians. At each Gaussian center, the corresponding triplane feature is queried and decoded into Gaussian attributes.
}
\label{fig:arch}
\end{figure*}

\section{Method}

\ourmethod infers the parameters of pre-defined FLAME~\cite{li2017learning} from a single image.
The key insight is that the pre-trained prior from \lrm can be adapted to aid the prediction of compact parametric models.
To this end, we integrate the analysis-by-synthesis pipeline for regressing FLAME into LRM's NeRF-targeting framework.

\cref{fig:ffhq} shows the overall architecture of \ourmethod, which consists of three key components: (1) a pre-trained image-to-triplane transformer from LRM,
(2) a 3D-aware decoder to predict FLAME parameters with the triplane tokens, and 
(3) a lightweight appearance decoder to convert the radiance-relevant features to surface-bounded 2D Gaussian primitives.
\ourmethod makes use of the image-to-triplane transformer to predict the feature-level triplane representation $\mathcal{T}$ of the input image $\mathbf{I}$.
Based on the triplane-based 3D representation, we predict FLAME parameters and model the local appearance with surface-bounded Gaussians $\{G\}$, so that we can render efficiently and thus become able to use 2D supervision in the image space to supervise the network.

\subsection{Transforming Images as Triplanes}

\lrm employs an image-to-triplane (I2T) transformer to lift 2D image features into the canonical 3D space as triplane representations.
The I2T transformer is designed with minimal inductive bias and pre-trained on large-scale datasets, enabling it to capture rich geometric and appearance priors.
It first extracts visual features by DINOv2~\cite{oquab2024dinov2}, and then camera intrinsic and extrinsic parameters $\mathbf{c} \in \mathbb{R}^{20}$ are flattened, concatenated, and embedded to modulate learnable positional embeddings (denoted as PE in \cref{fig:arch}). 
The final triplane tokens are upsampled to a higher spatial resolution and organized as the triplane representation used for synthesis.

During training, we freeze the pre-trained I2T transformer and keep the camera embedder trainable.
This design preserves the representations learned by the I2T transformer, while allowing the model to adapt to the viewpoint distribution of the facial dataset.
In early training, facial geometry is still inaccurate, and unreliable supervision may corrupt the learned tri-plane features.
Moreover, as \ourmethod fine-tunes \lrm on a new face dataset with viewpoints differing from LRM pre-training, freezing the camera embedder would limit effective adaptation.

\subsection{Predicting FLAME with Triplane Tokens}

\lrm decodes triplane tokens into a triplane-based radiance and occupancy field, from which a surface can be extracted by applying marching cubes~\cite{lorensen1987marching}.
This implies that the triplane tokens encapsulate sufficient geometric information, motivating \ourmethod to instead decode them into the parametric space of FLAME.

To effectively decode triplane tokens into FLAME parameters, we introduce a self-gating mechanism to selectively emphasize the token representation. 
As different facial geometries activate different subsets of triplane tokens, it is crucial to adaptively identify geometry-relevant tokens for each input image.
To this end, we treat the flattened tokens as a sequence and apply a lightweight gating module.
A shallow MLP followed by a sigmoid activation assigns a learnable importance score to each token, which is used to modulate the token features by suppressing geometry-irrelevant responses.

We follow DECA~\cite{feng2021learning} in predicting the same set of FLAME parameters: shape parameters $\boldsymbol{\beta} \in \mathbb{R}^{100}$, expression parameters $\boldsymbol{\psi} \in \mathbb{R}^{50}$, and pose parameters $\boldsymbol{\theta} \in \mathbb{R}^{6}$, which include global head pose and jaw pose.
Unlike DECA~\cite{feng2021learning}, which estimates 2D translation under a scaled orthographic camera model, we instead predict a global scaling factor $s \in \mathbb{R}_+$ and a translation vector $\mathbf{t} \in \mathbb{R}^3$ in world coordinates, enabling novel view synthesis beyond the input view.
The FLAME mesh $\mathcal{M}$ is reconstructed as
\begin{equation}
    \mathcal{M} = s \cdot V(\boldsymbol{\beta}, \boldsymbol{\psi}, \boldsymbol{\theta}) + \mathbf{t},
\end{equation}
where $V$ is the FLAME model detailed in supplementary.

\subsection{Synthesizing Appearance by Gaussians}

Since we explicitly predict FLAME parameters and reconstruct a FLAME mesh, we no longer require ray marching to repeatedly query triplane features, as used in the radiance field rendering of \lrm.
We therefore replace volume rendering with a rasterization-based approach using Gaussian Splatting~\cite{kerbl20233d} to improve rendering efficiency.
Among Gaussian-based representations, we adopt 2D Gaussian Splatting (2DGS)~\cite{huang20242d}, as it enables surface-aligned appearance modeling and provides more accurate depth and normal estimates for geometric supervision.

A straightforward way to bind 2D Gaussian primitives to the FLAME mesh is to initialize Gaussian centers at mesh vertices.
However, as noted in GaussianAvatars~\cite{qian2024gaussianavatars}, the $5,023$ vertices of a FLAME mesh are insufficient to capture the realistic appearance of the full head.
To address this, we sample $8k$ points on the FLAME mesh via differentiable barycentric interpolation, ensuring dense coverage of the facial surface.
For each 3D point $x\in \mathbb{R}^3$ sampled from the FLAME mesh $M$, we query the corresponding triplane feature $\mathcal{T}(x)$ and decode it with an MLP into Gaussian attributes, including opacity, scale, rotation, and color.
We apply different activation functions to different Gaussian attributes.
Following LGM~\cite{tang2024lgm}, the scale parameters are activated using SoftPlus and further scaled by a factor of $0.1$.
Opacity is activated with a standard sigmoid function, while colors are predicted using a slightly saturated sigmoid $f(x) = (1+2\epsilon)/(1+\exp(-x)) - \epsilon$ with $\epsilon\!=\!0.001$, as used in Mip-NeRF~\cite{barron2021mip}.
The predicted quaternions are $\ell_2$-normalized to ensure unit norm.

Once the 2D Gaussians are obtained, novel views can be synthesized by splatting them under arbitrary cameras intrinsics and extrinsics.
This enables photometric supervision during training, effectively encouraging the predicted 3DMM to capture fine-grained facial expressions and details.

\subsection{Loss Function}

We integrate an analysis-by-synthesis pipeline for FLAME estimation into the pre-trained LRM.
Since this integration introduces new supervision signals and optimization requirements, we adapt LRM's loss functions by modifying existing terms and introducing additional ones.

\paragraph{Photometric Loss.}
The photometric loss encourages the appearance consistency between the rendered and input images, and consists of three terms:
\begin{equation} \mathcal{L}_{\textrm{photometric}}=\mathcal{L}_{\textrm{pixel}}+\mathcal{L}_{\textrm{D-SSIM}}+\mathcal{L}_{\textrm{VGG}}.
\end{equation}
The pixel loss $\mathcal{L}_{\textrm{pixel}}$ encourages pixel-wise color consistency in facial regions, and is defined as
\begin{equation}
    \mathcal{L}_{\textrm{pixel}}=\|(\textbf{I}_{\textrm{target}}-\textbf{I}_\textrm{{render}})\odot(\lambda \facemask+(1-\lambda)\mathbf{1})\|^2,
\end{equation}
where $\odot$ denotes Hadamard product and $\facemask$ stands for the pixel-wise binary mask for facial regions.
$\lambda$ balances the influence of facial and non-facial regions.
The perceptual loss~\cite{dosovitskiy2016generating,johnson2016perceptual} and D-SSIM loss are computed for the entire image to encourage perceptual consistency.

\paragraph{Geometric Binding Loss.}
To encourage a tight coupling between the FLAME
mesh and 2D Gaussian primitives, we employ a binding loss that minimizes the depth and normal difference between 2DGS and FLAME:
\begin{multline}
    \mathcal{L}_{\textrm{binding}}=\|(\mathbf{D}_{\textrm{mesh}}-\mathbf{D}_\textrm{{gs}})\odot \facemask\| \\
    + \|(\mathbf{N}_{\textrm{mesh}}-\mathbf{N}_\textrm{{gs}})\odot \facemask \|,
\end{multline}
where $\mathbf{D}_{\textrm{mesh}}$ and $\mathbf{N}_{\textrm{mesh}}$ denote differentiably rendered depth and normal maps from the FLAME mesh, while $\mathbf{D}_{\textrm{gs}}$ and $\mathbf{N}_{\textrm{gs}}$ are alpha-blended ones according to Gaussian attributes.
Because 2D Gaussian primitives are highly expressive, facial appearance can be reconstructed without faithfully capturing the underlying geometry.
The binding loss is therefore essential to prevent the model from overfitting to appearance while neglecting geometric accuracy. 

\paragraph{Geometric Regularization Terms.}
As a common practice in FLAME regression, we also compute the reprojection error $\mathcal{L}_\textrm{lmk}$ of the 68 landmarks.
We further apply regularization terms 
$\|\boldsymbol{\beta}\|^2$ and $\|\boldsymbol{\psi}\|^2$, to prevent FLAME parameters from leaning to bizarre shapes and expressions (\eg, extremely long ears or a protruding forehead), as is a common practice~\cite{li2017learning, feng2021learning}.

\paragraph{Staged Training.}
Based on empirical observations, we use a two-stage training strategy for \ourmethod.
The first stage (stage I) trains \ourmethod only with landmark supervision $\mathcal{L}_\textrm{lmk}$ and regularization on shape and expression parameters $\mathcal{L}_\textrm{reg}$, facilitating the learning of head localization and orientation.
This stage reduces the optimization difficulty and degrees of freedom, providing a stable initialization for subsequent geometric refinement.
In the second stage (stage II), photometric and binding losses are further introduced to refine the reconstruction, encouraging both visual fidelity and geometric consistency.
All loss terms are applied per view over all source views, except for the view-independent shape and expression regularization terms.

%% file: sec/4_experiment.tex
\section{Experiments}

\subsection{Datasets}

We follow a rendering protocol similar to that of Objaverse~\cite{deitke2023objaverse}, as is used in the pre-training of OpenLRM~\cite{openlrm}, and apply it to textured 3D human head datasets, including FaceScape~\cite{zhu2023facescape}, Multiface~\cite{wuu2022multiface}, FaceVerse~\cite{wang2022faceverse}, and Headspace~\cite{dai2020statistical}.
The only modification is that, rather than uniformly sampling viewpoints over the full sphere, we restrict sampling to the frontal hemisphere to avoid uninformative rear-view renderings.
For each 3D asset, we normalize the mesh to fit within the bounding box of $[-1,+1]^3$ and render 32 random views, with all cameras sharing identical intrinsics and pointing toward the origin. 
To further enhance the performance in in-the-wild scenarios, we finetune \ourmethod on two in-the-wild monocular portrait datasets, FFHQ~\cite{karras2019style} and CelebA~\cite{liu2015deep}.
We evaluate the performance on the test sets of FaceVerse~\cite{wang2022faceverse} and FFHQ~\cite{karras2019style}, and benchmark shape accuracy on the NoW dataset~\cite{sanyal2019learning}.

\begin{table}[t]
\centering
\renewcommand{\arraystretch}{0.9}
\setlength{\tabcolsep}{4pt}
{
\footnotesize
\begin{tabular}{l|ccc|cc|cc}
\toprule
\multirow{2}{*}{\textbf{Methods}} & \multicolumn{3}{c|}{\textbfit{Chamfer Distance}} &  \multicolumn{2}{c|}{\textbfit{Shape $\textrm{Var}(\beta)$}} & \multicolumn{2}{c}{\textbfit{Expr. $\textrm{Var}(\psi)$}} \\
& mean & median & std & 10 & full & 10 & full \\
\midrule
DECA~\cite{feng2021learning} & 3.17 & 3.13 & 0.331 & ~~1.91 & ~~2.02 & ~~2.32 & ~~2.48 \\
EMOCA~\cite{danvevcek2022emoca} & 3.15 & 3.11 & 0.359 & ~~1.91 & ~~2.01 & ~~4.61 & ~~4.67 \\
SMIRK~\cite{retsinas20243d} & 3.20 & 3.18 & 0.329 & ~~9.47 & ~~8.48 & 45.94 & 47.04 \\
\textcolor{gray}{MICA~\cite{zielonka2022towards}} & \textcolor{gray}{2.63} & \textcolor{gray}{2.58} & \textcolor{gray}{0.381} & \textcolor{gray}{27.41} & \textcolor{gray}{28.52} & \textcolor{gray}{--} & \textcolor{gray}{--} \\
\midrule
\textbf{Ours} & \textbf{2.68} & \textbf{2.67} & \textbf{0.157} & \textbf{~~1.46} & \textbf{~~1.54} & \textbf{~~1.01} & \textbf{~~1.10} \\
\bottomrule
\end{tabular}
}
\caption{Quantitative comparison on FaceVerse test dataset~\cite{wang2022faceverse}. We report Chamfer distances ($\times 10^{-2}$) and cross-view variances of FLAME parameters predicted from single-view inputs under different viewpoints; lower is better for both metrics. 
``Full'' denotes the variance computed over the full parameter dimensions ($100$ for shape and $50$ for expression), while ``$10$" refers to the first 10 dimensions of each.
}
\label{tab:faceverse_variance}
\end{table}

\subsection{Implementation Details}

The implementation is based on the open-source code of OpenLRM~\cite{openlrm}, which is implemented by PyTorch, with mixed precision powered by Accelerate~\cite{accelerate}.
The differentiable renderer for FLAME depth and normal maps is implemented by PyTorch3D~\cite{ravi2020pytorch3d}.
We train our model using 4 NVIDIA A100 80GB SXM GPUs, with input images resized to $224 \times 224$ and rendered images at a resolution of $192 \times 192$.
Training on multi-view datasets takes approximately 4 days, while fine-tuning on monocular portrait datasets takes approximately 6 days.
The weighting factor $\lambda$ is set to 0.7 to place greater emphasis on facial regions.

For the multi-view datasets, we render $4$ source views per subject in one feedforward pass, with a batch size of 16 subjects per GPU.
The facial masks $\facemask$ used in stage II are obtained by projecting the face-wise semantic masks of FLAME~\cite{li2017learning} onto the image plane.
We generate ground-truth 3D landmarks for the datasets not providing.
To ensure multi-view consistency of landmarks, we first detect 2D landmarks by FAN~\cite{bulat2017far}, triangulate them by the 5 most frontal views to minimize the reprojection error, and then project the 3D landmarks to different views.
Note that the triangulated landmarks could still be error-prone so the loss weight should be carefully tuned.

For fine-tuning on monocular datasets, we render a single input view per subject and set the batch size to $32$ per GPU.
We employ off-the-shelf segmentation methods~\cite{kvanchiani2023easyportrait,kirillov2023segment} to obtain facial masks, and use FAN~\cite{bulat2017far} to detect 68 facial landmarks~\cite{sagonas2013300}.
To resolve the inherent scale–depth ambiguity in monocular reconstruction, (\eg, an object appearing larger or closer), we fix the $z$-component of the predicted translation vector to zero during training.

\begin{figure}
    \centering
    \includegraphics[width=\linewidth]{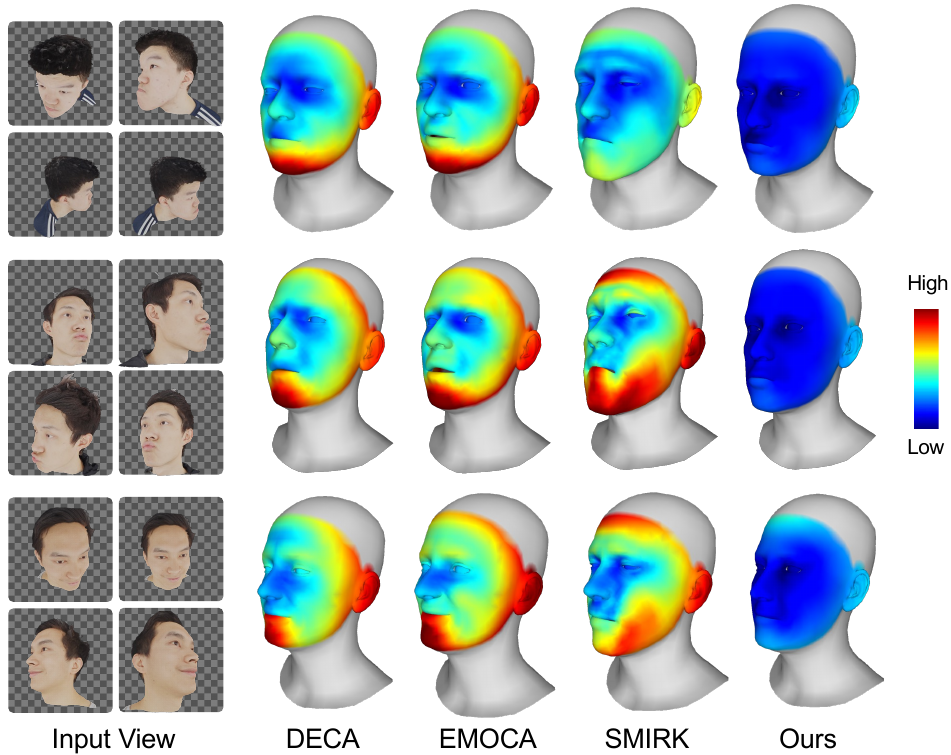}
    \caption{Per-vertex variances in local geometries. The heatmap is plotted on the mean face of multi-view predictions after keypoint alignment. Ours achieves more consistent predictions across viewpoints.}
    \label{fig:variance_compare}
\end{figure}

\subsection{Evaluation Protocol}

To demonstrate the effectiveness of our method, we conduct a twofold evaluation: the absolute accuracy of the reconstructed geometry, and the relative consistency of FLAME predictions across different viewpoints.

\begin{figure*}[t]
\centering
\includegraphics[width=\linewidth]{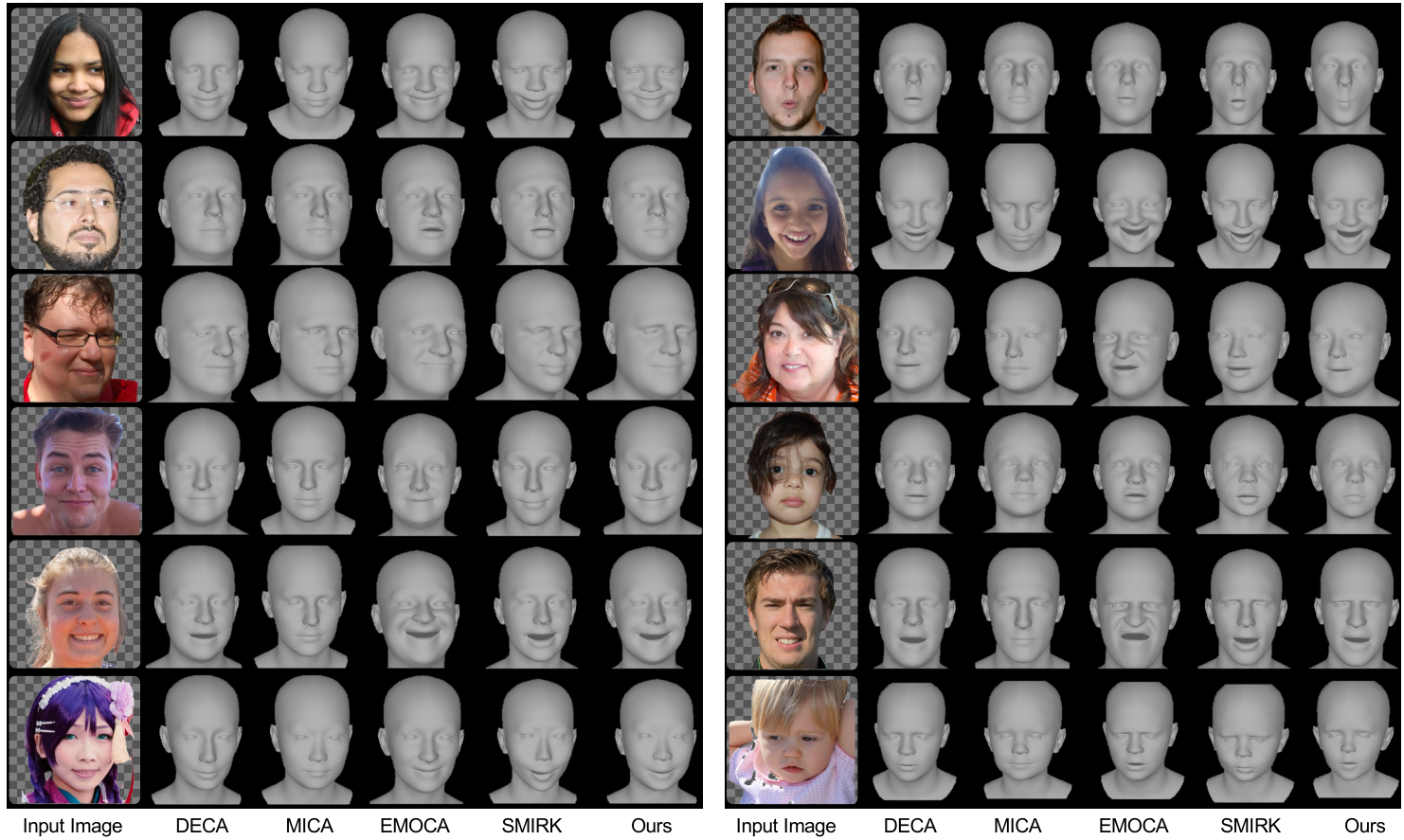}
\caption{Qualitative comparison on the test set of FFHQ dataset~\cite{karras2019style}. We align the results of different methods via an estimated transformation of the keypoints. 
Ours reconstructs shapes more accurately and recovers expressions more faithfully.}
\label{fig:ffhq}
\end{figure*}

For quantitative comparisons on FaceVerse~\cite{wang2022faceverse}, we follow the default evaluation protocol of NoW~\cite{sanyal2019learning}, computing the Chamfer distance between the predicted mesh and the ground truth over cropped facial regions after performing Umeyama alignment on the estimated keypoints.
For geometric accuracy on the NoW benchmark~\cite{sanyal2019learning}, we adopt the default non-metrical evaluation protocol. 
Following the benchmark setting, expression parameters and poses are set to zero, as only the neutral face geometry is evaluated.

As noted in \cite{wang2023sunstage}, a good 3DMM regressor should produce consistent predictions across multi-view inputs. 
Since \ourmethod leverages LRM’s prior for 3D-aware FLAME regression, it is expected to achieve strong consistency and robustness. 
To verify this, we additionally evaluate the variance of the predicted FLAME parameters under different camera views.
Let $N$ denote the number of views of the same subject. 
The variance of the shape parameters is
\begin{equation}
    \textrm{Var}(\boldsymbol{\beta})=\frac{1}{N}\sum_{i=1}^{N} \bigl\lVert \boldsymbol{\beta}_i - \overline{\boldsymbol{\beta}}\bigl\rVert_2^2,\; \text{with} \; \overline{\boldsymbol{\beta}} = \frac{1}{N}\sum_{i=1}^{N} \boldsymbol{\beta}_i.
\end{equation}
The variance for the expression parameters $\textrm{Var}(\boldsymbol{\psi})$ is defined in the same manner.

\paragraph{Baselines.}
We compare \ourmethod with \textbf{(1)} analysis-by-synthesis methods trained with 2D image supervision, including DECA~\cite{feng2021learning}, EMOCA (v2)~\cite{danvevcek2022emoca}, and SMIRK~\cite{retsinas20243d}, and \textbf{(2)} methods trained with direct FLAME parameter supervision, represented by MICA~\cite{zielonka2022towards}.

\subsection{Results}

\paragraph{FaceVerse Dataset.}
The quantitative results of the absolute geometric accuracy and the variance of multi-view predictions on FaceVerse~\cite{wang2022faceverse} are shown in 
\cref{tab:faceverse_variance}.
Since the FLAME parameters, $\boldsymbol{\beta}$ and $\boldsymbol{\psi}$, are obtained via Principal Component Analysis (PCA), the contribution of each parameter decreases from lower to higher indices. 
To provide more detailed insight, we additionally report variations of the first 10 components.
Overall, \ourmethod outperforms the baselines trained with 2D supervision in both metrics.

\Cref{tab:faceverse_variance} reveals that SMIRK~\cite{retsinas20243d} exhibits considerable sensitivity to viewpoint variations, particularly evident from the high variance of its predicted expression parameters. 
Similarly, MICA~\cite{zielonka2022towards} shows noticeable sensitivity to view changes in shape prediction.
As MICA relies on linear probing over ArcFace features~\cite{deng2020sub}, it primarily maps 2D facial appearance features to the FLAME parameter space, making it more sensitive to visible appearance cues.
This observation is consistent with our hypothesis that prior methods largely depend on 2D visual features for FLAME prediction.
\Cref{fig:variance_compare} further visualizes geometric variance by color coding per-vertex variances, where \ourmethod exhibits improved cross-view consistency.

\paragraph{FFHQ Dataset.}

As multi-view datasets are typically collected under highly controlled conditions, they often lack ethnic diversity and contain a limited number of samples.
To improve the performance, we further fine-tune \ourmethod on monocular portrait datasets. 
Since FFHQ dataset does not provide ground-truth 3D meshes, we conduct a qualitative comparison.
As shown in \cref{fig:ffhq}, \ourmethod demonstrates perceptually accurate reconstruction of overall facial shape while faithfully recovering facial expressions.
In contrast, DECA~\cite{feng2021learning} and SMIRK~\cite{retsinas20243d} often fail to capture accurate facial shape, while EMOCA~\cite{danvevcek2022emoca} and SMIRK~\cite{retsinas20243d} tend to predict exaggerated expressions.

\begin{table*}[t]
\centering
\renewcommand{\arraystretch}{0.9}
\resizebox{0.95\linewidth}{!}{
{\footnotesize
\begin{tabular}{l|cccccccc|gg}
\toprule
\multirow{2}{*}{\textbf{Methods}} & \multicolumn{2}{c}{\textbfit{Neutral}} & \multicolumn{2}{c}{\textbfit{Expressions}} & \multicolumn{2}{c}{\textbfit{Occlusions}} & \multicolumn{2}{c|}{\textbfit{Selfie}} & \multicolumn{2}{g}{\textbfit{Overall}}\\
& mean & median & mean & median & mean & median & mean & median & mean & median\\
\midrule
FLAME-neutral~\cite{li2017learning} & 1.36 &	1.70 &	1.38 &	1.73 &	1.39 &	1.77 &	1.35 &	1.72 & 1.37 &	1.73\\
DECA~\cite{feng2021learning} & 1.20 &	1.52 &	1.20 &	1.50 &	1.34 &	1.66 &	1.28 &	1.62 & 1.24 &	1.56\\
EMOCA~\cite{danvevcek2022emoca} & 1.19 &	1.47 &	1.16 &	1.45 &	1.28 &	1.55 &	1.25 &	1.56 & 1.21 &	1.49\\
SMIRK~\cite{retsinas20243d} & 1.00 &	1.24 &	1.02 &	1.25 & \textbf{1.03} &	\textbf{1.24} &	1.04 &	1.27 & 1.02 &	1.25\\
\textcolor{gray}{MICA~\cite{zielonka2022towards}} & \textcolor{gray}{0.96} & \textcolor{gray}{1.19} & \textcolor{gray}{0.95} & \textcolor{gray}{1.17} & \textcolor{gray}{0.96} & \textcolor{gray}{1.18} & \textcolor{gray}{1.00} & \textcolor{gray}{1.24} & \textcolor{gray}{0.96} & \textcolor{gray}{1.19} \\                 
\midrule
\textbf{Ours} & \textbf{0.98} &	\textbf{1.22} &	\textbf{0.98} &	\textbf{1.20} &	\textbf{1.03} &	1.26 &	\textbf{1.01} &	\textbf{1.24} & \textbf{0.99} &	\textbf{1.23} \\
\bottomrule
\end{tabular}
}}
\caption{Challenge-wise and overall prediction errors, reported as Chamfer distances in $mm$ on the NoW validation set~\cite{sanyal2019learning}.
}
\label{tab:now}
\end{table*}

\paragraph{NoW Benchmark.}
We further evaluate the shape accuracy of all methods on the validation set of the NoW benchmark~\cite{sanyal2019learning}. 
While the benchmark nominally assesses only the shape parameters $\boldsymbol{\beta}$, the monocular estimation of FLAME shape and expression is intrinsically ill-posed. 
Consequently, the NoW evaluation also serves as an indirect measure of a method’s ability to decouple shape from expression.
The quantitative results are summarized in \cref{tab:now}, where Chamfer distances are reported across different challenge categories.
Notably, \ourmethod is trained on substantially fewer samples and with less identity diversity than competing approaches.
For example, DECA~\cite{feng2021learning} is trained on $\sim$2 million in-the-wild portrait images, while our model is trained on only $\sim$0.3 million.
This validates the effectiveness of leveraging the pre-trained prior of LRM learned from general objects for parametric facial reconstruction.

\subsection{Ablation Study}

We conduct an ablation study to evaluate the impact of training-related design choices, as summarized in \cref{tab:ablation}.
Each component is found to contribute positively to geometric reconstruction quality.

First, removing the binding loss results in the poorest reconstruction quality, as evidenced by the largest Chamfer distance, highlighting the necessity of enforcing a tight coupling between the FLAME mesh and the 2D Gaussian primitives.
Notably, removing the binding loss also yields the most consistent cross-view FLAME predictions.
This is because, without explicit geometric constraints, the highly expressive 2D Gaussian representation can absorb cross-view appearance variations, reducing the need for the model to explain photometric differences through changes in FLAME geometry. As a result, the predicted FLAME parameters vary less across viewpoints, despite degraded geometric accuracy.

Second, photometric supervision is essential for high-fidelity geometric reconstruction.
Without the photometric supervision, \ourmethod relies primarily on the landmark loss, which alone is insufficient to recover detailed facial geometry.
Finally, we attempt to train \ourmethod without initializing LRM with pre-trained parameters.
In this setting, training fails to converge, likely because the available multi-view facial images are insufficient to train LRM from scratch.
As capturing large-scale multi-view facial data is laborious, pre-training LRM on general objects becomes crucial for reducing the data requirements of \ourmethod.

\begin{table}[t]
\centering
\renewcommand{\arraystretch}{0.9}
{\footnotesize
\begin{tabular}{l|ccccc}
\toprule
\multirow{2}{*}{\textbf{Settings}} &  \multicolumn{3}{c}{\textbfit{Chamfer Distance}} & \multicolumn{2}{c}{\textbfit{Variance}}  \\
&  mean & median & std & $\boldsymbol{\beta}$ & $\boldsymbol{\psi}$ \\
\midrule
\textit{w/o} regularization & 2.75 & 2.74 & 0.170 & 2.08 & 0.41\\
\textit{w/o} photometric loss & 2.99 & 2.96 & 0.200 & 2.07 & 0.41\\
\textit{w/o} binding loss & 3.05 & 3.05 & 0.206 & \textbf{0.75} & \textbf{0.27}\\
\textit{w/o} pre-training & \multicolumn{5}{c}{--} \\
\midrule
\textbf{Full} & \textbf{2.68} & \textbf{2.67} & \textbf{0.157} & 1.54 & 1.10 \\
\bottomrule
\end{tabular}
}
\caption{Ablation study on the test set of FaceVerse dataset~\cite{wang2022faceverse}. 
Variance is computed over the full parameter dimensions.
Under the setting of ``\textit{w/o} pre-training'', the network fails to produce meaningful outputs, and we leave it blank.
}
\label{tab:ablation}
\end{table}

%% file: sec/5_discussion.tex
\subsection{Limitations}
Compared to the closed-source Rodin~\cite{wang2023rodin} dataset comprising 100$k$ head objects, the amount of training 3D assets available to us remains relatively limited.
In addition, Gaussian-based representations typically require a high degree of redundancy to achieve photorealistic rendering.
In contrast, our method employs significantly fewer Gaussians than approaches optimized purely for rendering quality~\cite{qian2024gaussianavatars}, as perfect visual fidelity is not our primary objective.
Furthermore, we observe that our method is less effective at capturing eyelid and eyeball movements.
This limitation is not reflected in the Chamfer distance, as open and closed eyes exhibit similar local geometry.
We attribute this issue to the fact that, due to privacy concerns, eye regions of facial images are often removed or masked out in many 3D head datasets, resulting in limited photometric supervision for these areas.

%% file: sec/6_conclusion.tex
\section{Conclusion}
We have presented \ourmethod, a framework that adapts the pre-trained prior of LRM to parametric face reconstruction in an analysis-by-synthesis fashion. 
By projecting pre-trained triplane features into the FLAME parameter space and coupling the FLAME mesh with surface-bounded 2D Gaussian primitives for appearance modeling, \ourmethod effectively bridges implicit 3D representations with interpretable parametric models. 
Experiments demonstrate that our method achieves competitive reconstruction accuracy and improved cross-view consistency compared to existing approaches, even when trained with fewer samples and less identity diversity. 
Our results indicate that the multi-view knowledge and generic 3D representations learned by LRM on general objects can facilitate parametric face estimation when multi-view training data is limited.

%% file: sec/X_suppl.tex
\clearpage
\setcounter{page}{1}
\maketitlesupplementary

In this supplementary document, we provide preliminary knowledge for our method (\cref{sec:prelimi}), extra discussions for a better understanding of our approach and results (\cref{sec:dis}), additional visual results (\cref{sec:visual_res}), and a declaration of the sources of assets (\cref{sec:assets}).

\section{Preliminaries}
\label{sec:prelimi}

\subsection{FLAME}

FLAME~\cite{li2017learning} is a statistical 3D head model that combines separate linear identity shape and expression spaces with linear blend skinning (LBS) and pose-dependent corrective blendshapes to articulate the neck, jaw, and eyeballs.
It can be described as a function mapping of $\mathbb{R}^{|\beta|\times |\psi|\times |\theta|} \to \mathbb{R}^{3\times 5023}$ from the latent parameters to the coordinates of mesh vertices as 
\begin{equation}
    V(\boldsymbol{\beta},\boldsymbol{\psi},\boldsymbol{\theta}) = W(T_P(\boldsymbol{\beta},\boldsymbol{\psi},\boldsymbol{\theta}),J(\boldsymbol{\beta}),\boldsymbol{\theta}, \mathcal{W}),
\end{equation}
where $W(\mathbf{T},J,\boldsymbol{\theta},\mathcal{W})$ denotes a standard skinning function to rotate the vertices of $\mathbf{T}$ around joints $J$ by blending weights $\mathcal{W}$.
Further, we have
\begin{equation}
    T_P(\boldsymbol{\beta},\boldsymbol{\psi},\boldsymbol{\theta}) = \overline{\mathbf{T}} + B_S(\boldsymbol{\beta};\mathcal{B}) + B_P(\boldsymbol{\theta};\mathcal{P}) + B_E(\boldsymbol{\psi};\mathcal{E}),
\end{equation}
where $B_S$, $B_P$, and $B_P$ are blendshape functions for shape, pose, and expression, with weights $\mathcal{B}$, $\mathcal{P}$, and $\mathcal{E}$ respectively.
$\overline{\mathbf{T}}$ represents a template mesh under ``zero pose."
As a result, the mesh reconstruction $\mathcal{M}$ is
\begin{equation}
    \mathcal{M} = s\cdot V(\boldsymbol{\beta},\boldsymbol{\psi},\boldsymbol{\theta}) + \mathbf{t},
\end{equation}
where a global scaling $s$ and translation $\mathbf{t}$ is applied.

Note that in \ourmethod, we apply FLAME as a pre-trained decoder for geometry and it is possible to replace FLAME with other parametric models.

\subsection{2D Gaussian Splatting}

In \ourmethod, we opt to model the avatar's appearance via 2D Gaussian Splatting~\cite{huang20242d}, where a point is represented as an oriented 2D Gaussian disk parameterized by the disk center $\boldsymbol{\mu}\in\mathbb{R}^3$, two principal tangent directions $\mathbf{t}_u,\mathbf{t}_v \in \mathbb{R}^3$ and corresponding scales $s_u, s_v \in \mathbb{R}_+$.
Then the $uv$-parametrization of the plane where 2D Gaussian lies is
\begin{equation}
    P(u,v) = \boldsymbol{\mu}+s_u\mathbf{t}_uu + s_v\mathbf{t}_vv = \mathbf{H}[u,v,1,1]^{\top},
\end{equation}
where $\mathbf{H}\in \mathbb{R}^{4\times 4}$ is the parameterized plane 
\begin{equation}
    \mathbf{H} = \begin{bmatrix}
    \mathbf{R}\mathbf{S} & \boldsymbol{\mu} \\
    0 & 1
    \end{bmatrix}\,,
\end{equation}
where the rotation matrix $\mathbf{R}\in SO(3)$ can be deduced by $\mathbf{t}_u$ and $\mathbf{t}_v$, and the scale matrix $\mathbf{S}=\textrm{diag}(s_u,s_v,0)$.
In practice, we maintain the quaternion representation $\mathbf{q} \in \mathbb{R}^4$ for the rotation instead of the tangent directions.
In \ourmethod, we do not estimate an offset for each Gaussian for a better 2DGS-mesh-alignment, so we have $\boldsymbol{\mu}=\textbf{x}$.

Different from 3DGS~\cite{kerbl20233d}, 2DGS adopts a ray-splat intersection to prevent the numerical instability of degenerated cases.
It allows 2DGS to compute depth maps in a more precise way and normal maps in a faster closed-form fashion, than the original 3DGS, making 2DGS more suitable for geometric modeling.
In \ourmethod, we opt for 2DGS for its better alignment between appearance and geometry since we are targeting FLAME estimation.

\section{Extra Discussions}
\label{sec:dis}

Given the limitation on the number of pages, we have provided a concise discussion in the main body, with more detailed information available in the supplementary material.

\paragraph{Forms of Supervision.}
In the main paper, we have categorized the methods for FLAME prediction as 2D supervised (\eg, DECA~\cite{feng2021learning}, EMOCA~\cite{danvevcek2022emoca}, and SMIRK~\cite{retsinas20243d}), and 3D supervised (\eg, MICA~\cite{zielonka2022towards}).
We would like to emphasize that such categorization could be considered rather coarse. 
A more precise distinction lies in whether the supervision is provided directly or indirectly (via analysis by synthesis). 
Direct supervision is generally easier for networks to learn from, while indirect supervision requires the network to acquire additional knowledge that does not directly participate in evaluation but still significantly influences the results, \eg, the global scale and translation.

It's also worth noting that, during the pre-training process of SMIRK~\cite{retsinas20243d}, the shape prediction by MICA~\cite{zielonka2022towards} is used for supervision.

\paragraph{FoV Issue.}
When finetuning \ourmethod with CelebA dataset~\cite{liu2015deep} and FFHQ dataset~\cite{karras2019style}, which are in-the-wild portrait datasets, we have to pre-set a field of view (FoV) for \ourmethod to re-render the reconstructed avatars.
As is pointed out in CameraHMR~\cite{patel2025camerahmr}, the perspective effect plays an important role in the final reconstruction quality.

We therefore test 3 different choices of FoV for synthesizing the in-the-wild datasets, namely 73.43$^{\circ}$, 14.25$^{\circ}$, and 2.87$^{\circ}$.
Given a square image, the relationship between the field of view (FoV) and the image's side length $L$ is
\begin{equation}
    \textrm{FoV} = 2\cdot \textrm{arctan}\left(\frac{L}{2f}\right) = 2\cdot\textrm{arctan}\left(\frac{1}{2\alpha}\right),
\end{equation}
where $f=\alpha L$.
Therefore, we have tested the cases in \cref{tab:fov} with $\alpha = 0.75, 4.0, 20.0$, under which the perspective effect becomes progressively weaker.

\begin{table}[t]
\centering
\resizebox{0.75\linewidth}{!}{
\renewcommand{\arraystretch}{0.9}
{
\footnotesize
\begin{tabular}{cccc}
\toprule
$\alpha$ & FoV & mean & median \\
\midrule
0.75 & 73.43$^{\circ}$	& 1.02 & 1.26\\
4.0 & 14.25$^{\circ}$ & 1.00 & 1.24\\       
20.0 &  2.87$^{\circ}$ & 0.99 & 1.23\\
\bottomrule
\end{tabular}
}
}
\caption{Quantitative evaluation of the effect by varying FoV at training time. The reported metrics are Chamfer distances on the validation set of NoW~\cite{sanyal2019learning}.}
\label{tab:fov}
\end{table}

\section{More Visual Results}
\label{sec:visual_res}

\begin{figure*}[t]
    \centering
    \includegraphics[width=0.9\linewidth]{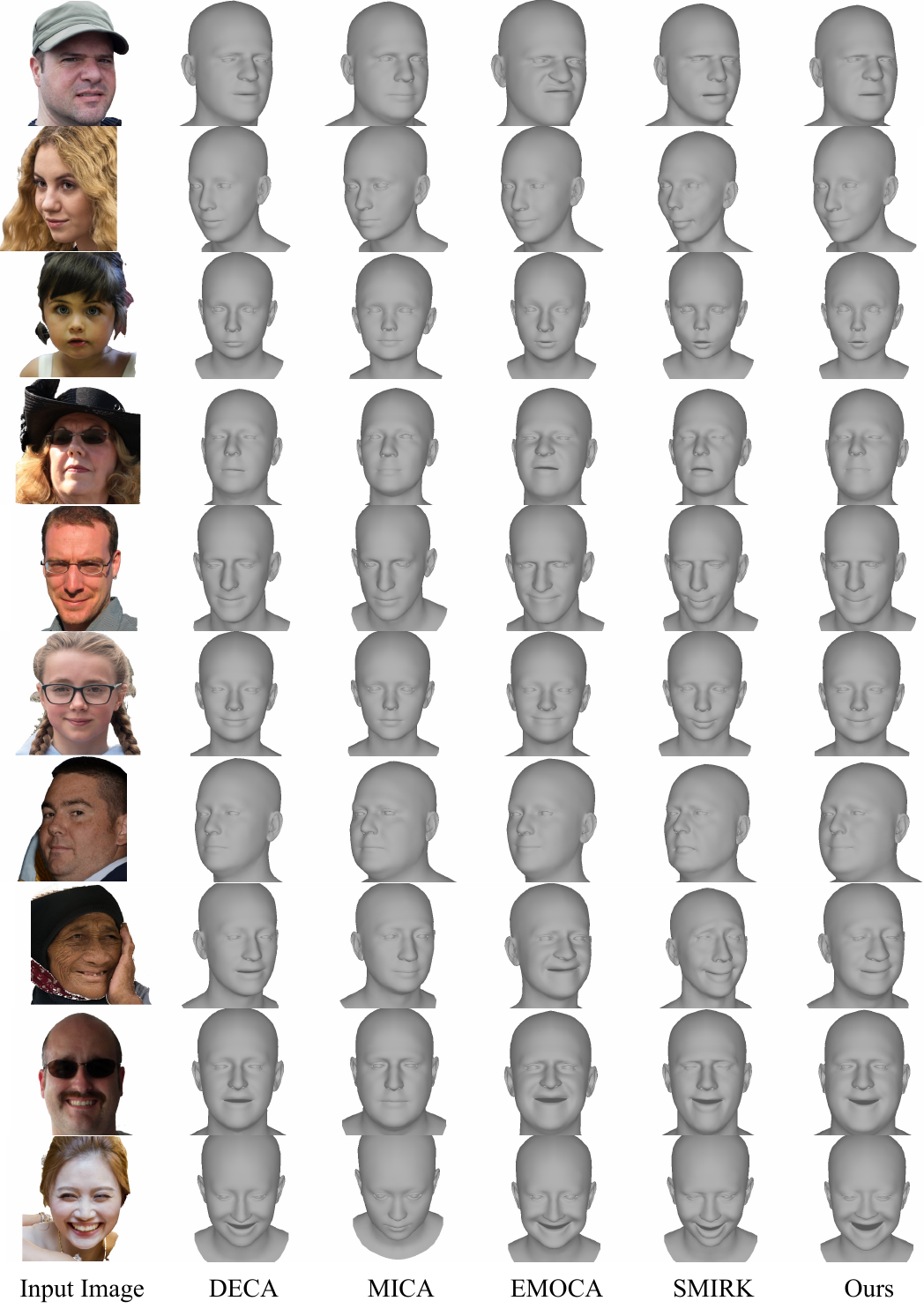}
    \caption{More qualitative results on the test set of FFHQ~\cite{karras2019style}.}
    \label{fig:ffhq_supp}
\end{figure*}

\paragraph{FFHQ Results.}
We demonstrate more visualized comparisons on the test set of FFHQ dataset~\cite{karras2019style} in \cref{fig:ffhq_supp}.
In line with the main paper, we encourage readers to pay attention to both the global head geometry and the detailed facial expressions.

\paragraph{Video Results.}
In the supplementary material, we include a short video that illustrates the variations in the predicted FLAME meshes under different viewpoints, demonstrating that \ourmethod achieves relatively better consistency.
The video further includes the FLAME tracking results on video sequences from VFHQ~\cite{xie2022vfhq}, providing an intuitive illustration of temporal consistency.

\section{Use of Existing Assets}
\label{sec:assets}

We here list all the existing assets used in this manuscript and would like to sincerely appreciate the maintainers of these open-source projects.
\begin{itemize}
    \item OpenLRM~\cite{openlrm}: \url{https://github.com/3DTopia/OpenLRM}
    \item 2D Gaussian Splatting~\cite{huang20242d}: \url{https://github.com/hbb1/2d-gaussian-splatting}
    \item PyTorch3D~\cite{ravi2020pytorch3d}: \url{https://github.com/facebookresearch/pytorch3d}
    \item Blender: \url{https://projects.blender.org/blender/blender}
    \item FaceVerse dataset~\cite{wang2022faceverse}: \url{https://github.com/LizhenWangT/FaceVerse-Dataset}
    \item FaceScape dataset~\cite{zhu2023facescape}: \url{https://facescape.nju.edu.cn}
    \item Multiface dataset~\cite{wuu2022multiface}: \url{https://github.com/facebookresearch/multiface}
    \item Headspace dataset~\cite{dai2020statistical}: \url{https://www-users.york.ac.uk/~np7/research/Headspace}
    \item CelebA dataset~\cite{liu2015deep}: \url{https://mmlab.ie.cuhk.edu.hk/projects/CelebA.html}
    \item FFHQ dataset~\cite{karras2019style}: \url{https://github.com/NVlabs/ffhq-dataset}
    \item NoW dataset~\cite{sanyal2019learning}: \url{https://now.is.tue.mpg.de}
\end{itemize}